# Impoved RPN for Single Targets Detection based on the Anchor Mask Net

Mingjie Li[1], Youqian Feng[1], Zhonghai Yin[1], Cheng Zhou[1], Fanghao Dong[1]

[1]Air Force Engineering University, Foundation Department, Xi'an,China

**Abstract.** Common target detection is usually based on single frame images, which is vulnerable to affected by the similar targets in the image and not applicable to video images. In this paper , anchor mask is proposed to add the prior knowledge for target detection and an anchor mask net is designed to impove the RPN performance for single target detection. Tested in the VOT2016, the model perform better.

**Keywords:** Anchor mask; RPN; Single targets detection; Timing; Time series

## 1. Introduction

Since the introduction of deep learning technology into the field of computer vision, all tasks for single-frame images are done well by a variety of networks. In the field of image recognition, Alexnet , VGG, Resnet[1] and more constantly refreshed the correct rate record of the Imagenet game. In the field of Target segmentation, FCN[2] and perfect application of the technology of CRF[3][4] on it make the segmentation effect is more and more significant. AND in the field of Target Detection, the series of RCNN[5][6][7], the series of YOLO[8][9][10], R-FCN[11] and SSD[12] perform better and better by constantly improving the structure. But the target detection on the singe image frame can be affected simply if there is no prior knowledge added in the network. And with the development of target detection networks, the basic structure of RPN in also used in the all mainstream algorithm(for instance Faster-RCNN ,Yolo-v3, and SSD). According to the problem and character of RPN , the concept of anchor mask is proposed in this paper, which can be used to connect to the prior knowledge to filter a lot of wrong anchors to improve the detection accuracy. An anchor mask net considering the timing characteristic is designed to join to the RPN as an auxiliary part. Passing the test in the vot2016 data set , new RPN preforms better for single targets detection.

## 2. Relation work

### 2.1 RPN

RPN is proposed firstly in Faster-RCNN, which generate several anchor boxes with different scales and ratios according to the anchor points on feature maps. And anchor boxes can be adjusted to location the target and provided confidence scores by CNN. RPN is a effective structure to get proposal boxes which replace the complex procession of research boxes generation in RCNN. The structure of RPN is as showed in Figure 1.

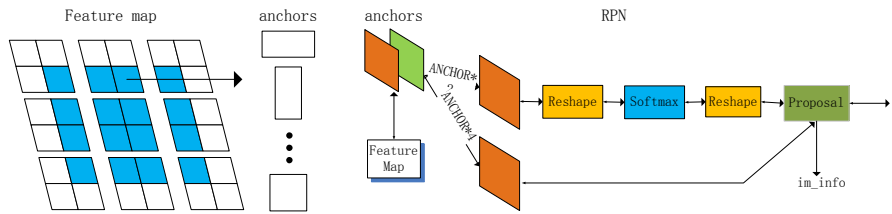

Figure 1

In this paper, a RPN is trained to be a target detector for video clip frames. Because RPN is the basic structure for the main steam algorithm in the targets detection field. So our anchor mask net will be useful in other networks if it is fine to RPN.

### 2.2 FCN

FCN is a classical structure for target segmentation. It gets the image features by the convolutional layers and pool layer firstly. After that, resize the feature maps to the same size of the original image by the fully convolutional layers and de-convolutional layers to predict the classes of each pixel.

In this paper, a similar FCN is trained to predict the valid anchor points. The structure of FCN is shown in Figure 2.

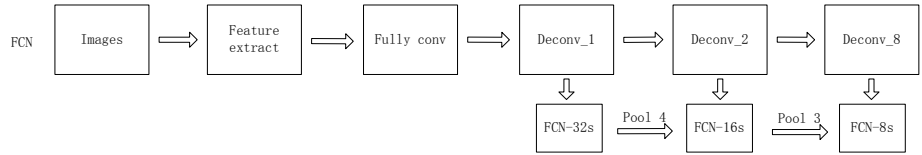

Figure 2

## 2.3 3dCNN

Three-dimensional Convolutional neural Networks(3dCNN) is firstly use for video analysis proposed in the C3D[13]. This structure integrates temporal and spatio information of video. Similar to ordinary convolution operations, .3dCNN make clip frames at different times equivalent to channels in 2dcnn. Convolutional operation is performed in each time dimension, and then an addition operation is performed on all time channels. In this paper, a 3dCNN is used to be the front end to get time information of the IOU heat maps of the first three frames. The principle of 3dCNN is shown in Figure 3.

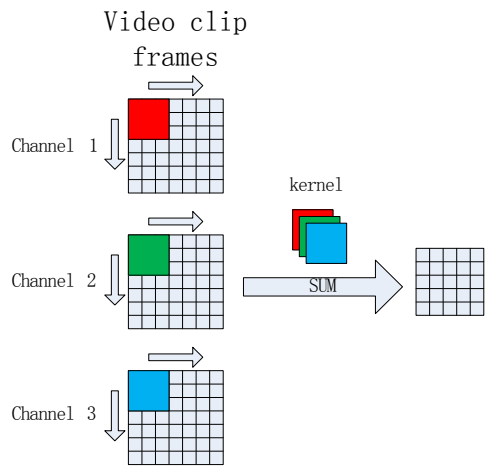

Figure 3

## 3. RPN with anchor mask net

### 3.1 anchor mask and IOU heat map

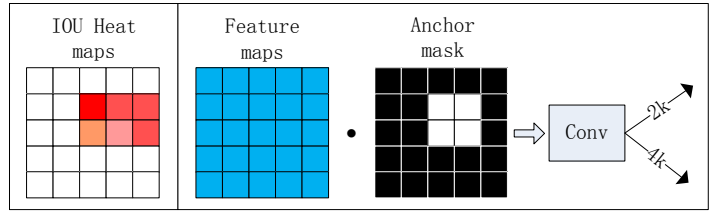

Figure 4

In RPN, all the anchor points on the feature maps will be gotten operation to predict a box data and confidence scores. But there is always a few boxes marked valid, and the anchor boxes of interference target reserved. In this paper, a matrix containing only 0 and 1 is proposed to be the filter to dot multiply with feature maps before the convolution operation to get position and scores (The procession is shown in Figure 4).The matrix which is called anchor mask is a filter to integrate prior knowledge and eliminate impossible locations. According to the highest score of all the anchor of per anchor points when we make the result boxes as ground truth, a heat map with same two-dimensional size as feature maps is made. In the heat mop, the value of each pixel is the highest IOU value of the anchors of the corresponding anchor point, and set the value less than threshold to 0. A heat map is show in Figure 4.

### 3.2 RPN with anchor mask net

In order to consider the prior knowledge of the video of the previous frame and excluded interference target, a RPN with anchor mask net is designed in this paper. The structure of network is shown in Figure 5.

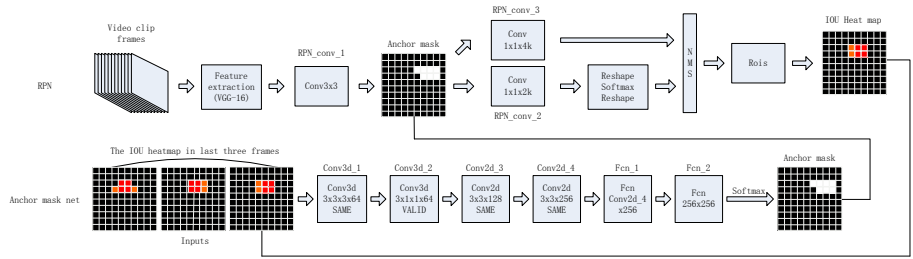

Figure 5

There are two stage in this network. And two network are trained in two stage. In stage 1, the IOU heat maps of last three frames was sent to the network. After two 3d Conv layers, input is compressed into one-dimensional. And after two 2d Conv layers to extract features and two Fully Convolutional layers and one Softmax layer, an anchor mask of next images which pixel is prediction of confidence scores generates.

In stage 2, the anchor mask generating in stage 1 dot multiply the feature maps before the generation of proposal boxes. According to rois of Non-maximal suppression(nms), the IOU heat map of this image generates which is used to predict anchor mask of next image. In this paper , the front end of vgg-16 is used to be the feature extractor and all the video clips are resize to $(224, 224)$.

**3.3 losses of network**

The two networks in two stages in this be trained separately. In this paper, the video clip frames are resized to $(224, 224)$ before put into RPN. And losses of RPN are same as Faster-RCNN. In this paper, the RPN is trained without anchor-mask to increase negative samples and the ratio between negative and positive sample is 3(according to the SSD). The labels of others negative sample is set to -1 meaning invalid. Losses of the RPN is as Formulas 1,2 and 3.

$$loss_{rpn\_total} = loss_{rpn\_scores} + loss_{rpn\_reg} \quad (1)$$

$$loss_{rpn\_scores} = \frac{-\sum_{i}^{n} labels_i \cdot \log(socres_i)}{n} \quad (2)$$

$$loss_{rpn\_reg} = \frac{\sum_{i}^{n}\sum_{j}^{4} R(t_j - t_{j^*})}{n} \quad (3)$$

Where $loss_{rpn\_total}$ is the total loss of RPN include the classification loss $loss_{rpn\_score}$ and regression loss $loss_{rpn\_reg}$ in Formulas 1. Where n is the number of valid anchors and $labels_i$ is the truth classification of $i-th$ anchor in Formulas 2. Where $t_j$ is one of $(t_x, t_y, t_h, t_w)$ which is the corresponding parameters between the ground truth and anchors and $(t_{x^*}, t_{y^*}, t_{h^*}, t_{w^*})$ is the corresponding parameters between the predict boxes and anchors. $R$ is Robust loss function $smooth_{L1}$.

And the anchor mask net is trained with the IOU heat maps of the last threes clip frames. Losses of anchor mask net is shown in Formula 4.

$$loss_{mask} = \frac{-\sum_{i}^{M} label\_p_i \cdot \log(score\_p_i)}{M} \quad (4)$$

Where M is the number of anchor points in the feature maps. $label\_p$ is the ground truth anchor mask and $score\_p$ is the confidence scores of each pixel of anchor mask.

## 4. Experiment

### 4.1 Experimental environment and train losses

Hardware Environment: AMD RYZEN 2200 and GTX1080-Ti.
Software Environment: Ubuntu16.04+tensorflow+pycharm

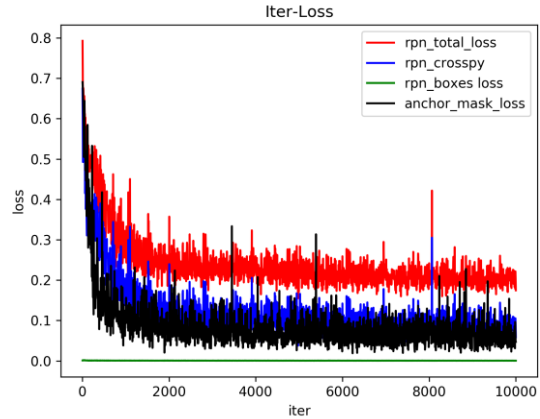

Our model get trained in the vot2016 which is a common target tracking data set. The data set consists of short video slices of 60 different objects. In this paper, 20 images are randomly selected as a batch to train RPN. And 20 randomly selected 3 clip frames(or less) are used to train Anchor mask net as a batch. All models were trained 10,000 times, and training losses were recorded every 5 times. The relationship of Iteration and loss is shown in Figure 6.

As the Figure 6 shown, all losses have dropped significantly over time. And the loss of anchor mask net drops quickly. It means that the anchor mask net is useful to get temporal characteristic. The loss of RPN dropping shows RPN is useful to detect target.

### 4.2 Result of anchor mask net

RPN and anchor mask net are trained separately and used together as Figure5. The model is tested on vot2016 data set in continuous video clip frames. The part of result of test is shown in Figure 8.

The blue boxes are ground truth boxes. The red boxes are results of RPN with anchor mask net. And the green boxes are results of RPN.

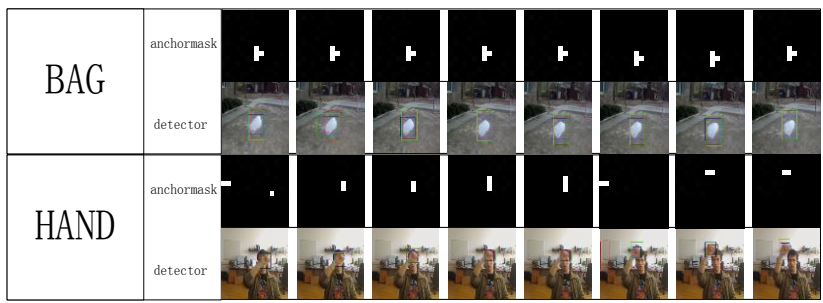

Figure 8

As the Figure 8 shown, the IOU of our model is higher than ordinary RPN. But it doesn't work well if the prediction of anchor mask is wrong. (For example, the result of sixth images of hand is wrong because of the wrong prediction of anchor mask).So the model is improved in the paper.

**4.3 The improved model of RPN with anchor mask**

For the problem of wrong prediction of anchor mask, a new generation method of anchor mask is proposed. The IOU heat map of last frame is added to the prediction to increase detection range. The new anchor mask comprehensive considerate of recent pictures and predicted information, which can be used to fix bug prediction results. The result of new RPN is    shown in Figure 9.

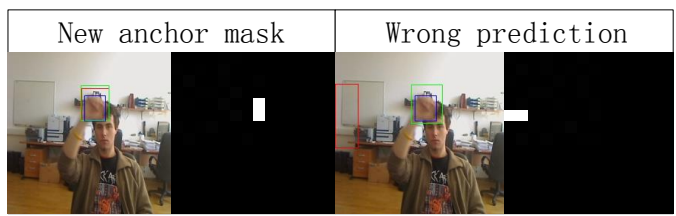

Figure 9

New anchor mask is useful as Figure 9 shown, our model is useful for single target detection of video clip frames.

## 5. Conclusion

A RPN with anchor mask net is proposed in this paper, which considerate the spatio and temporal characteristic. The model is tested and improved in the vot2016 data set, the effect of model is ideal.